%% file: main.tex
\documentclass[journal]{IEEEtran}

\usepackage{cite}
\usepackage{hyperref}

\usepackage[pdftex]{graphicx}

\usepackage{amsmath}
\usepackage{amsfonts}

\usepackage{booktabs}

\usepackage[caption=false,font=normalsize,labelfont=sf,textfont=sf]{subfig}
\usepackage{todonotes}
\usepackage{xcolor}
\definecolor{rev}{RGB}{77, 117, 117}

\usepackage{soul}

\usepackage{makecell}

\usepackage{fixltx2e}

\usepackage{float}

\usepackage{url}

\begin{document}
%

\title{Authorship Impersonation via LLM Prompting does not Evade Authorship Verification Methods}

%
%

\author{Baoyi Zeng, Andrea Nini}  

%
%

\markboth{March~2026}%
{Impersonation of LLM}
%



\maketitle

%
\IEEEpeerreviewmaketitle

\input{abstract}
\input{article}
\input{appendix}
\ifCLASSOPTIONcaptionsoff
  \newpage
\fi

\bibliographystyle{IEEEtran}

\bibliography{references}

\end{document}

%% file: abstract.tex
\begin{abstract}

Authorship verification (AV), the task of determining whether a questioned text was written by a specific individual, is a critical part of forensic linguistics. 
While manual authorial impersonation by perpetrators has long been a recognized threat in historical forensic cases, recent advances in large language models (LLMs) raise new challenges, as adversaries may exploit these tools to impersonate another's writing. 
This study investigates whether prompted LLMs can generate convincing authorial impersonations and whether such outputs can evade existing forensic AV systems. 
Using GPT-4o as the adversary model, we generated impersonation texts under four prompting conditions across three genres: emails, text messages, and social media posts. We then evaluated these outputs against both non-neural AV methods (n-gram tracing, Ranking-Based Impostors Method, LambdaG) and neural approaches (AdHominem, LUAR, STAR) within a likelihood-ratio framework. 
Results show that LLM-generated texts failed to sufficiently replicate authorial individuality to bypass established AV systems. We also observed that some methods achieved even higher accuracy when rejecting impersonation texts compared to genuine negative samples. 
Overall, these findings indicate that, despite the accessibility of LLMs, current AV systems remain robust against entry-level impersonation attempts across multiple genres.
Furthermore, we demonstrate that this counter-intuitive resilience stems, at least in part, from the higher lexical diversity and entropy inherent in LLM-generated texts.
    
\end{abstract}


%% file: article.tex
\section{Introduction}
\IEEEPARstart{A}{uthorship} 
verification (AV) is a long-standing challenge in forensic linguistics and digital forensics. 
One of the core tasks of AV is to determine, given two or more texts, whether they were produced by the same author, and the outcome may contribute to evidence considered in court for disputed authorship. 
Typically, an analyst  receives a set of \emph{known} texts 
$\mathbb{D}_\mathcal{A} = \{\mathcal{D}_1, \mathcal{D}_2, \dots\}$ authored by an individual $\mathcal{A}$, 
alongside one or more \emph{unknown} texts $\mathcal{D}_\mathcal{U}$ of disputed authorship $\mathcal{U}$. 
The goal is to decide whether $\mathcal{A} = \mathcal{U}$, i.e., whether the unknown text $\mathcal{D}_\mathcal{U}$ was indeed written by $\mathcal{A}$~\cite{stamatatos2009survey, halvani2021practice, koppel2012fundamental}.

However, AV systems are increasingly confronted with adversarial attacks, in which a writer deliberately manipulates their text to influence attribution outcomes. 
Two prominent forms of such attacks are \emph{impersonation} and \emph{obfuscation}~\cite{mcdonald2012use, altakrori2022evaluation, kacmarcik2006obfuscating, alperin2025masks}. 
In impersonation, the adversary attempts to mimic the linguistic characteristics of a target author, 
to ensure that their text is falsely attributed to that individual. 
In obfuscation, by contrast, the goal is to conceal one’s own authorship, for example, by suppressing idiosyncratic markers or introducing noise, so that attribution to the true author becomes difficult or impossible. 
Both strategies pose serious challenges to forensic analysis and can be exploited by malicious actors to confound investigations.

With the rapid advancement of large language models (LLMs), these systems have demonstrated strong capabilities across a wide range of natural language processing tasks, including text style transfer (TST), where texts are rewritten to approximate the linguistic characteristics of a target domain or individual~\cite{fu2018style, sudhakar2019transforming}. 
Such mimicry can be achieved through various techniques, including prompting~\cite{mikros2025beyond, liu2024adaptive}, parameter-efficient fine-tuning~\cite{mukherjee2024large, reif2021recipe}, and memory-augmented models\cite{toshevska2024large} (for a review,  see~\cite{toshevska2025llm}). 
Recent studies further demonstrate that LLMs can imitate target authors in ways that often remain undetected by human readers, 
highlighting their potential for authorial impersonation~\cite{shi2025impersona, herbold2024large}.
Moreover, these models are now widely accessible to the public through commercial interfaces,  developer APIs, and open-source releases.  

This accessibility raises the possibility that malicious actors could exploit LLMs to generate fabricated texts that mimic the linguistic identity of a target author, thereby complicating authorship attribution and undermining the reliability of linguistic analysis as forensic evidence.
Historical cases in Forensic Authorship Analysis demonstrate the practical implications of impersonation.
In both the murders of Danielle Jones in 2001 and Jenny Nicholl in 2005, the perpetrators used the victims' mobile phones to text their families in order to mislead the investigations~\cite{coulthardIntroductionForensicLinguistics2017}. 
Similarly, in 2009, Christopher Birks murdered his wife Amanda and then used her phone to text himself and some of the victim's friends~\cite{grantTXT4N6Method2013}. 
In 2012, Jamie Starbuck murdered his wife Debbie while on a cruise around the world and then sent emails to her family and friends, pretending to be her for almost three years~\cite{grantStarbuckCaseMethods2022}. 
Notably, Jamie Starbuck noticed how Debbie frequently used a semi-colon in her emails and reproduced this feature of her language while impersonating her. 
This attempt at impersonation was identified by the forensic linguists who worked on this case, who were able to tell that Debbie used the semi-colon with a different syntactic function than Jamie. 
Although these manual impersonation attempts temporarily fooled some of the recipients, they were not sophisticated enough to pass a detailed linguistic analysis performed by a linguist.  
However, the advent of LLMs introduces unprecedented capabilities that may complicate similar forensic investigations in the future.

While the forensic community has long acknowledged the risk of attempts to disguise or manipulate writing style~\cite{kacmarcik2006obfuscating, brennan2009practical}, the rise of LLMs represents a qualitatively different threat. 
This development parallels the proliferation of ``deepfakes'' in audio and video, which have profoundly disrupted digital forensics by challenging the authenticity of multimedia evidence~\cite{qureshi2024deepfake, chadha2021deepfake}. 
Just as audiovisual deepfakes synthesise highly realistic but fabricated identities, LLMs now facilitate the creation of ``textual deepfakes'' that convincingly mimic an individual's authorial voice.
Earlier adversarial strategies relied on manual or algorithmic editing, often employing rule-based methods~\cite{castro2017author}. 
Such approaches were time-consuming and frequently degraded the fluency or semantic similarity between the resulting text and the original text~\cite{crothers2022adversarial}.
By contrast, LLMs can generate stylistically controlled texts at scale, requiring minimal effort or expertise from the user~\cite{shokri2025personalized}. 
This ease of authorship impersonation amplifies the potential forensic impact, yet, systematic evaluations of LLM impersonation against established AV methods remain scarce.

Although a growing body of work has investigated LLMs for text style transfer and stylistic imitation, most of this research has been conducted within NLP benchmarking and evaluation contexts rather than applied to forensic settings (e.g.,~\cite{hu2022text, hu2025learning, jin2022deep}). 
This paper therefore addresses this gap by simulating a realistic adversarial scenario: a malicious actor, relying solely on publicly available resources, uses prompting to produce impersonation texts and a forensic linguist must decide whether those texts were written by the claimed author. 
We focus on prompting because it is the most accessible and available attack vector. While more advanced adversaries might attempt parameter-level adaptation (e.g., fine-tuning or personalized model training), such approaches require substantial data and resources, placing them beyond the scope of the present study.

Our aim is to test whether prompted LLMs can successfully impersonate real authors effectively enough to evade detection by AV methods.
This goal can be examined from two complementary perspectives: the capability of LLMs to reproduce an idiolect, and the robustness of AV methods when confronted with such impersonations.
By situating the evaluation within a forensic framework, this study assesses the resilience of current AV techniques against emerging adversarial threats and provides empirical insight into the risks posed by LLM-assisted authorship manipulation.


\section{Related Work}
\label{sec: lr}

Research on prompting LLMs for text style transfer or impersonation has largely focused on well-known figures, such as literary authors or public personalities~\cite{mikros2025beyond, dinu2025analyzing, tao2024cat, herbold2024large}. 
While such studies demonstrate the models’ ability to approximate recognisable styles, they exhibit two limitations. 
First, the writings of famous individuals may have been included in the LLMs’ pre-training data, raising concerns about data contamination and memorisation.
Second, these scenarios do not reflect the types of impersonation most relevant to forensic contexts, where malicious actors are more likely to impersonate private individuals in everyday communication~\cite{orebaugh2014visualizing, sainio2024brand}.

Moving beyond the imitation of high-profile figures, Yang and Carpuat~\cite{yang2025steering} used Biber’s Multi-Dimensional Analysis (MDA) to guide LLM prompting for arbitrary text style transfer. 
They employed similarity scores from an AV system (LUAR) to evaluate stylistic alignment and demonstrated that register-based prompting improves controllability and fidelity. 
However, their study utilised AV scores primarily as an evaluation metric for stylistic similarity, rather than assessing the generated texts as adversarial attacks designed to deceive established AV systems.

In contrast, other researchers have explicitly tested impersonation ability of LLMs as authorship attribution or verification problems. 
Bhandarkar et al.~\cite{bhandarkar2024emulating} examined whether off-the-shelf LLMs can emulate authorial features with minimal intervention, finding that models can capture surface-level traits but struggle with deeper stylistic signatures. 
Building on this, Chen and Moscholios~\cite{chen2024using} proposed a Complex Emulation Protocol that integrates author exemplars with linguistically informed instructions. Their results suggest that explicit linguistic cues improve stylistic alignment, though outcomes remain highly sensitive to prompt design.
Jemama and Kumar~\cite{jemama2025well} showed that under few-shot prompting conditions, LLMs can successfully imitate academic essay style, while Wang et al.~\cite{wang2025catch} found that models perform well on structured genres such as news or emails but struggle with informal registers like blogs and forums. 
Supporting this limitation in informal contexts, Thompson and Ishihara~\cite{chatgptReview_thompson2025unveiling} demonstrated that ChatGPT fails to impersonate individual styles in product reviews via one-shot learning. 
Instead of capturing personalised human markers, the model heavily reverts to its default, hyperbolic machine vocabulary.
Taken together, these findings indicate that impersonation performance varies substantially across genres and prompting strategies.

While previous studies suggest that LLMs are capable of authorship impersonation under certain conditions, their success remains heavily dependent on prompt design, genre, and evaluation criteria.
Crucially, prior work has predominantly assessed performance through linguistic similarity metrics or human judgments, rather than evaluating whether LLM-generated impersonations can evade the established AV systems deployed in forensic investigations.
Moreover, most studies treat AV as a classification task, without using the likelihood ratio framework that underpins the admissibility and interpretation of forensic evidence~\cite{morrison2012likelihood}.

Our study addresses these gaps by situating LLM impersonation explicitly within a forensic science framework. 
We systematically test whether prompted impersonations can bypass both non-neural and neural AV methods, and examine whether current verification paradigms remain resilient under adversarial conditions.

\section{Methods}
\label{sec: methods}

\subsection{Problem Formalisation}

We formalise the authorship impersonation task as follows. Let a corpus $\mathcal{C}$ (see Section~\ref{sec: corpora}) contain a set of authors $\mathbb{A} = \{\mathcal{A}_1, \mathcal{A}_2, \dots, \mathcal{A}_n\}$. 
For each author $\mathcal{A}_i$, we partition their texts into a training set and a test set; for methods that require validation, the training set is further divided into training and validation subsets. 
Given an author $\mathcal{A}_i$ with their known documents $\mathbb{D}_{A_i} = \{\mathcal{D}_1, \mathcal{D}_2, \dots\}$,
we provide $\mathbb{D}_{A_i}$ as examples to the LLM under a prompting strategy $P$ (see Section~\ref{sec: prompting}). 
Alongside these examples, the LLM is provided with a source text $\mathcal{S}$ written by a different author.
The model then produces an impersonation text ${\mathcal{Q}}_i^P$, by rewriting $\mathcal{S}$ to mimic the authorial identity of $\mathcal{A}_i$. 
This generated text ${\mathcal{Q}}_i^P$ then serves as the query text to be evaluated.

Each AV method $M$ (see Section~\ref{sec: eval}) takes as input the query text ${\mathcal{Q}}_i^P$ and the known texts $\mathbb{D}_{A_i}$ and produces a score $s_i^M$. 
These scores are then calibrated using a logistic regression calibration method~\cite{morrisonTutorialLogisticregressionCalibration2013}, yielding log-likelihood ratios (LLRs) that reflect the relative support for the same-author versus different-author hypotheses. 
Positive values indicate that the system judged the generated text as written by the target author and therefore that the method has been successfully deceived by the impersonation. Conversely, negative values indicate that the system judged the generated text as not written by the target author. 
Larger absolute values reflect a higher confidence in the decision, while values around zero signal an inconclusive result regarding authorship.

By analysing these calibrated scores across prompting strategies and verification methods, we can systematically assess both the impersonation capability of LLMs and the robustness of AV systems.

\subsection{Dataset}
\label{sec: corpora}

We conducted experiments on three corpora, representing distinct genres: emails, text messages, and social media posts (tweets). 
These genres are frequently encountered in forensic investigations~\cite{corney2003analysing}. 
This selection allows us to evaluate whether the effectiveness of LLM-based impersonation varies across domains with differing stylistic and structural properties. 
Importantly, these text types reflect realistic forensic scenarios: for example, a suspect might obtain a target’s historical emails or text messages and attempt to fabricate new material in that language, leaving forensic linguists to determine whether a questioned document was genuinely authored by the same individual who produced the known samples.
Table~\ref{tab:corpus_stats} summarises the descriptive statistics of the three corpora.

\begin{table}[H]
\centering
\caption{Descriptive statistics of the corpora used in the experiments.}
\label{tab:corpus_stats}
\begin{tabular}{ccccc}
\toprule
Corpus & \#Authors & \#Texts & \makecell{Avg. texts\\ per author} & \makecell{Avg. tokens\\ per author} \\
\midrule
$\mathcal{C}_\mathrm{Enron}$  & 76 & 345 & 4.54 & 243,618.30 \\
$\mathcal{C}_\mathrm{BOLT}$  & 46 & 3716 & 80.78 & 46,933.08 \\
$\mathcal{C}_\mathrm{Twitter}$ & 60 & 10933 & 182.22 & 150,984.73 \\
\bottomrule
\end{tabular}
\end{table}

\subsubsection{Enron Email Corpus}

The Enron Email Corpus\footnote{https://www.cs.cmu.edu/~enron/} consists of messages written by employees of the Enron Corporation and released during the investigation into the company’s collapse~\cite{klimt2004enron}. 
The collection originally contained data from approximately 150 users, most of whom were from senior management. 
The dataset provides valuable information about corporate communication, featuring relatively long texts and substantial variation in the amount of material available per author. 
Due to its scale, authenticity, and diversity, it has become one of the most widely used resources for research in authorship verification and related fields.

We used the version of the corpus preprocessed by Nini et al.~\cite{LambdaG-nini2024authorship}, in which URLs, email headers, signatures, and routine greeting or closing salutations were removed. 
The dataset was filtered to retain only authors with a sufficient amount of text. After preprocessing, the resulting subset comprised 345 texts written by 76 authors, forming the $\mathcal{C}_\mathrm{Enron}$ corpus used in our experiments.

\subsubsection{BOLT English SMS/Chat}

The BOLT English SMS/Chat Corpus\footnote{https://catalog.ldc.upenn.edu/LDC2018T19}, developed by the Linguistic Data Consortium (LDC), contains naturally occurring short message service (SMS) and online chat (CHT) data collected through both data donations and live interactions with native English speakers~\cite{song2014collecting}. 
Compared to the Enron corpus, BOLT texts are typically shorter, less carefully edited, and more interactive, thereby offering a distinct stylistic profile. This makes the corpus particularly valuable for testing whether impersonation attempts by LLMs generalise formal communication to informal conversational registers.

Preprocessing was applied to ensure textual consistency while preserving authorship signals. Non-ASCII characters (including emojis) were removed, word elongations were normalised, and URLs as well as non-printing object-replacement characters were stripped. As with Enron, only authors with sufficient material were retained for experimentation; in this case, we required a minimum of 1,000 tokens per author to ensure that verification results would be based on adequate data. We refer to this preprocessed version of the corpus as $\mathcal{C}_\mathrm{BOLT}$.

\subsubsection{Twitter}

We also included a Twitter corpus to extend our evaluation to the social media domain. This corpus was drawn from a large collection of geolocated tweets from users in the UK, originally collected via the Twitter API in 2014~\cite{twitterCorpus-grieve2019mapping}.

The Twitter data underwent preprocessing largely consistent with that applied to the BOLT corpus. Duplicate tweets within each author’s text set were removed. 
Texts were lowercased, and emojis, non-printing Unicode characters, leading punctuation, and redundant whitespace were stripped.
Word elongations were normalised, URLs were replaced with the placeholder \texttt{\textless url\textgreater}, and user mentions were anonymised as ``userid''. Finally, retweets beginning with ``rt :'' were excluded.

Total token counts were computed for each author, and authors with more than 135,000 tokens were selected as candidates. These accounts were manually inspected to remove non-human or automated sources (e.g. weather-report bots). After filtering, 60 authors with sufficient data remained. Among them, 30 authors were randomly assigned to the training set, with the remaining 30 used for the test set.
For each author, texts were randomly sampled to create a fixed-size subset comprising 2,000 tokens of known text and 500 tokens of unknown text. 
This design choice was made to ensure comparability in text length across the Twitter corpus and the other two corpora used in the study. The resulting collection constitutes the $\mathcal{C}_\mathrm{Twitter}$ corpus used in our experiments.

\subsection{Large Language Model}

All impersonation texts were generated using GPT-4o, a large language model developed by OpenAI~\cite{hurst2024gpt}. The model was accessed through the OpenAI API.

The model was chosen for three main reasons: 
(i)~it demonstrates strong performance across a wide range of tasks, including natural language generation, reasoning, and style transfer, and thus reflects a state-of-the-art LLM currently available to the public; 
(ii)~it has been benchmarked for stylistic control and in-context learning~\cite{chae2025large, tsai2021style, liu2021makes, wan2023gpt}, capabilities that are crucial for the task of authorship impersonation; 
and (iii)~it is readily accessible with minimal effort, thereby realistically simulating the conditions under which an adversary might attempt to deploy such a model.

\subsection{Source Text Selection}
\label{sec: source_text}

Because we frame the impersonation task as a rewriting exercise, the selection of the original source text ($\mathcal{S}$) is a crucial experimental design choice. 
For each corpus, we randomly selected one author to act as the source author. 
Notably, this source author was not used as a target author, ensuring that the model is never tasked with rewriting a text into the language of its original creator. 
We then used the unknown text of this source author as the global source text $\mathcal{S}$ for that specific dataset.

This selection strategy was employed for two main reasons. 
First, by drawing the source text from the same dataset as the target authors, we control for genre and domain distribution. 
Second, by using the same source text across all impersonation attempts within a given dataset, we hold the semantic content constant. Consequently, any variation in the verification scores ($s_i^M$) is strictly attributable to the LLM's ability to mimic the target author's language, rather than variations in the content of the source texts.

For $\mathcal{C}_\mathrm{BOLT}$ and $\mathcal{C}_\mathrm{Twitter}$, this choice resulted an unnatural source text made up of concatenated messages or tweets spanning various contexts.
However, our manual inspection of the outputs confirmed that the LLM had no difficulty processing these inputs, successfully producing impersonation texts that accurately retained the original content.

\subsection{Prompting Techniques}
\label{sec: prompting}

To probe whether LLMs can be prompted to impersonate a specific human author and evade AV systems, we employed four prompting conditions that vary in structure, explicitness, and the degree of prompt engineering. 
Specifically, all prompts framed the task as a rewriting exercise: the LLM was provided with text snippets from a target author and was asked to rewrite a distinct source text, producing the impersonation text.
All the prompts used in our experiment can be found in Appendix~\ref{app:prompts}.

\subsubsection{Naive Prompting}
This baseline models a minimally skilled adversary. 
The model is given a single-sentence instruction that asks it to rewrite a provided source text so that it resembles an example snippet from the target author to be impersonated.  
The model generates a single passage per trial without any self-reasoning, iterative planning, or additional guidance.

\subsubsection{Self-prompting}

We use a two-step self-prompting procedure to reduce reliance on manually crafted prompts and to simulate an attacker who specifies only a high-level goal. 
In the first step, the model is given a meta-prompt that states the rewriting goal and supplies examples from the target author; it is then asked to generate a detailed instruction for itself. 
The instruction typically takes the form of a checklist of features the model deems important for producing a convincing impersonation (e.g. lexical preferences and punctuation habits). 
Consequently, the model produces a fresh, author-specific prompt for each generation. Because different target authors exhibit different salient cues, this method allows those differences to be reflected in the instruction. In the second step, the model-authored instruction is fed back to the same model to produce the final text. 
This approach minimises bias from hand-engineered prompts and also reveals which linguistics cues are emphasised in the automatically generated instructions. 
Only the final generated passages are evaluated by the AV methods.

\subsubsection{Role-Playing Prompting}
We also implement a role-playing prompt by stacking a system prompt with a subsequent user prompt. 
In this setup, the system message first assigns a persistent role to the model (e.g.. a writing assistant specialising in mimicking linguistic characteristics). The user message then supplies the concrete task: the source text to be rewritten, exemplars from the target author, and a checklist of features to attend to. Separating role assignment from task specification keeps the impersonation objective active throughout generation, provides clearer control over the model’s behaviour, and lets us probe how explicit, checklist-style guidance affects impersonation success compared with single-message prompts.

In addition, we incorporate Linguistically Informed Prompting (LIP) following the approach of Huang et al.~\cite{huang2024can}. Their work showed that explicitly highlighting linguistic features in prompts improves LLM performance on authorship verification tasks. We therefore test whether the same strategy benefits impersonation generation. In practice, LIP directs the model to disregard the topic of the given examples and instead attend to features such as punctuation patterns, the use of phrasal verbs, the presence of rare words, and occasional errors.

\subsubsection{Tree-of-Thoughts Prompting}
ToT prompting combines structured multi-step generation with explicit linguistic guidance. The procedure follows the framework of Chen et al.~\cite{chen2024using} and divides each round of text generation into a planning stage and a drafting stage, with both stages incorporating candidate generation followed by voting.

In the planning stage, the model is given the task objective together with the examples and the target text. It is prompted to propose alternative plans for carrying out the impersonation task. To reduce cross-contamination between ideas, the model generates plans independently for each target author. For every impersonation attempt, the model produces three candidate plans and then conducts an internal vote across five rounds to select the best plan, which is the one most likely to succeed at the task. The winning plan is then carried forward to the next stage.

In the drafting stage, the model is instructed to follow the selected plan and generate five independent candidate passages. It then conducts a second internal vote to determine the best passage, which is retained as the final output. Only these final winning passages are evaluated by the AV algorithms.

\subsection{Evaluation}
\label{sec: eval}

To assess whether LLM-generated impersonations could evade detection, we evaluated them using a diverse set of established AV methods. 
The selection included both non-neural approaches, which are based on interpretable statistical measures or likelihood-based modelling, and neural network–based approaches, which make use of pre-trained language models and embedding representations.
These methods were chosen either because they are widely used in the authorship verification literature or because they are more recent and have been shown to outperform earlier approaches (\cite{nini2023theory, proisl2018delta, puspitasari2023identify}).

\subsubsection{Non-Neural Methods}
\paragraph{N-gram Tracing}
This method measures the similarity between documents based on frequent character or word n-grams\cite{ngramtracing-grieve2019attributing}. 
The underlying assumption is that short contiguous sequences of tokens encode the stable language use of authors, allowing for straightforward comparison without requiring external impostor documents. Although originally proposed as an authorship attribution method, n-gram tracing has more recently been tested for verification as well~\cite{nini2023theory}.
In our implementation, we used character 9-grams and computed similarity using the Simpson coefficient, following prior findings that this combination provides the best performance~\cite{nini2023theory}.

\paragraph{Ranking-Based Impostors (RBI) Method}

This method compares the questioned document with both the known author’s samples and a set of impostor texts authored by others~\cite{RBI-potha2017improved, potha2020improved}. 
RBI prioritises those impostor texts that are most similar to the known author’s texts, thereby forming a more competitive reference set. 
For each iteration, a random subset of features and impostors is selected, document similarities are computed, and the position (rank) of the known author’s text among the impostors determines the verification score. 
In our implementation, cosine similarity was used as the comparison coefficient, with parameters set to their default values ($k = 300$ features, $m = 100$ impostors, and $n = 25$ iterations). 

\paragraph{LambdaG}
Recently introduced by Nini et al.~\cite{LambdaG-nini2024authorship}, LambdaG is an AV method situated within the likelihood ratio paradigm of forensic science. The method constructs Grammar Models, which are n-gram language models estimated solely on function tokens after content-masking~\cite{halvani2021posnoise}. It then computes a log-likelihood ratio (called $\lambda_G$) between the candidate author’s grammar model and reference population models, quantifying whether the questioned text is more probable under the candidate author’s idiosyncratic grammar. Nini et al. \cite{LambdaG-nini2024authorship} show that this approach outperforms other established AV baselines.

\subsubsection{Neural Network-Based Methods}

\paragraph{AdHominem}

AdHominem (Attention-based Deep Hierarchical cOnvolutional siaMese bIdirectional recurreNt nEural-network Model) is an attention-based Siamese neural network designed for authorship verification~\cite{AdHominem-boenninghoff2019explainable}. It encodes texts hierarchically at the character, word, and sentence levels, before applying attention layers to highlight informative features. 
The model is trained with a metric learning objective so that documents by the same author are embedded closer together, while those by different authors are pushed apart in the latent space. 
By capturing fine-grained linguistic cues beyond surface lexical overlap, AdHominem provides a strong neural baseline for evaluating whether LLM-generated texts can mimic genuine authorship.

\paragraph{LUAR}
LUAR (Learning Universal Authorship Representations)~\cite{LUAR-rivera2021learning} builds on pre-trained sentence encoders and applies a supervised contrastive training objective to learn domain-transferable authorship embeddings. 
The model aggregates multiple short text segments per author and optimises their representations to ensure that texts by the same individual cluster tightly, while those from distinct authors are separated. 
Although LUAR is explicitly designed to capture authorial cues that are stable across topics, its performance still degrades significantly when subjected to substantial domain shifts.

\paragraph{STAR}
STAR (Style Transformer for Authorship Representations)~\cite{STAR-huertas2024understanding} is a transformer-based model that uses supervised contrastive pre-training to learn style-sensitive author representations. 
By optimising the embedding space to cluster texts by the same author while separating different authors, STAR effectively captures intrinsic linguistic features that go beyond surface-level lexical patterns.
Although originally evaluated in cross-topic social media settings, STAR is also well-suited for our experiments because its training strategy emphasises robust authorial features rather than surface-level lexical overlap. 
This makes it particularly relevant for testing whether LLM-generated texts can successfully imitate an author’s writing in controlled scenarios.

\subsection{Content-Masking}

Authorship verification models are often affected by topic bias, where classifiers may rely on semantic content rather than stylometric features. 
To mitigate this, Stamatatos~\cite{stamatatos2017authorship} proposed TextDistortion, which performs content-masking by replacing infrequent or topic-specific words with uniform symbols. While effective in cross-topic settings, this approach also removes potentially useful grammatical and functional elements.

Building on this idea, Halvani and Graner~\cite{halvani2021posnoise} introduced POSNoise, a preprocessing method designed to mask topic content more selectively. 
POSNoise replaces semantically rich tokens (such as nouns, verbs, adjectives, and adverbs) with their corresponding Part-of-Speech (POS) tags, while retaining function words, such as contractions and prepositions, as well as punctuation marks. 
This preserves more information associated with authorship, forcing verification systems to focus on linguistic patterns rather than meaning. Notably, the LambdaG method inherently requires this step.

In our experiments, we applied POSNoise preprocessing to all non-neural AV methods to minimise topic interference. However, since neural-based models rely on pre-trained embeddings that encode lexical semantics, POSNoise was not applied to these methods in order to avoid disrupting the learnt embedding space.

\section{Results}
\label{sec: results}

We begin by examining whether LLM-generated impersonations were able to bypass AV methods. 
To this end, we compared the LLR scores assigned by these methods to impersonation texts produced under four prompting conditions.

\subsection{Results of LLR}

\begin{figure}[ht]
    \centering
     \includegraphics[width=0.98\linewidth]{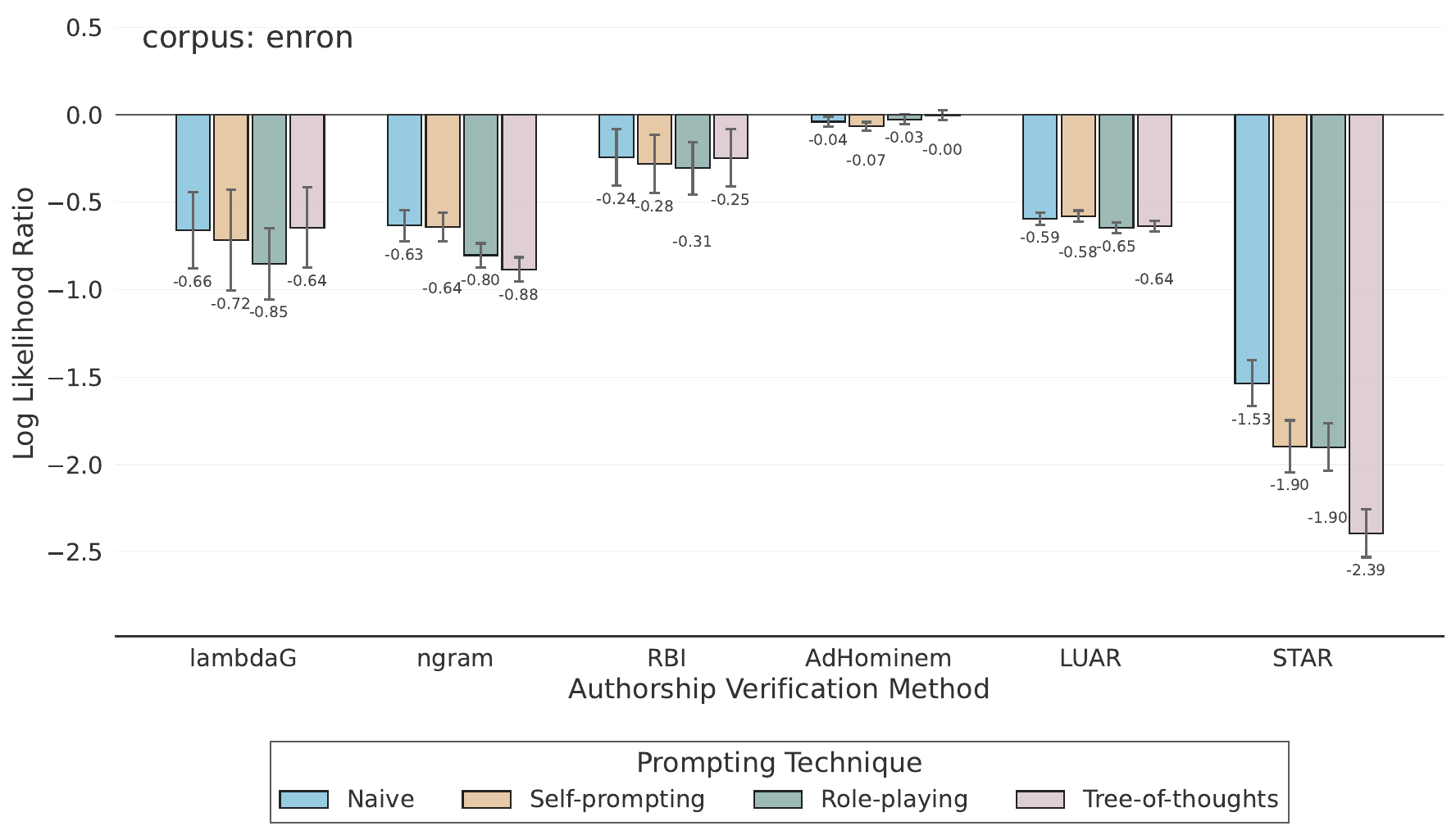}
     \includegraphics[width=0.98\linewidth]{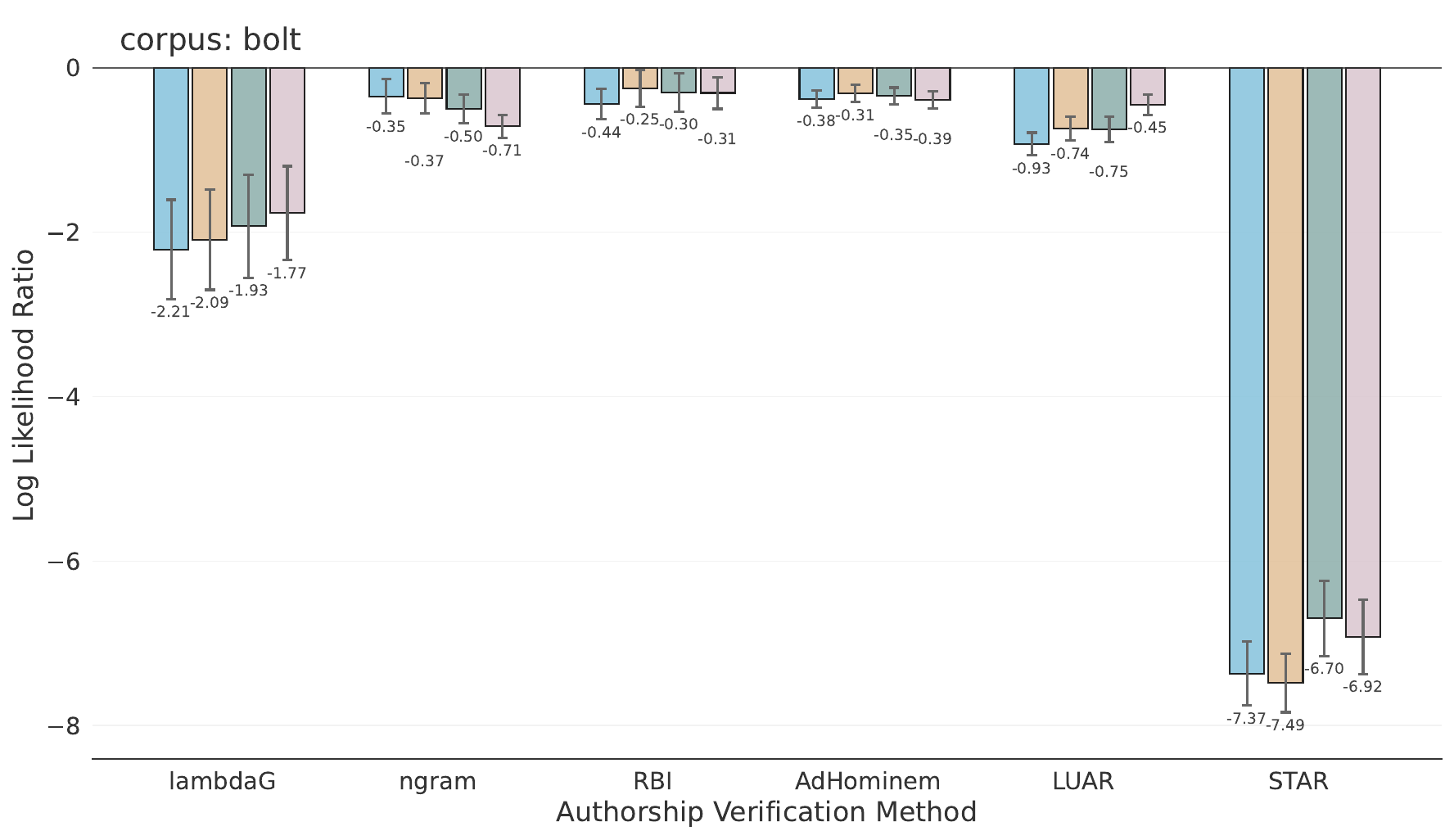}
     \includegraphics[width=0.98\linewidth]{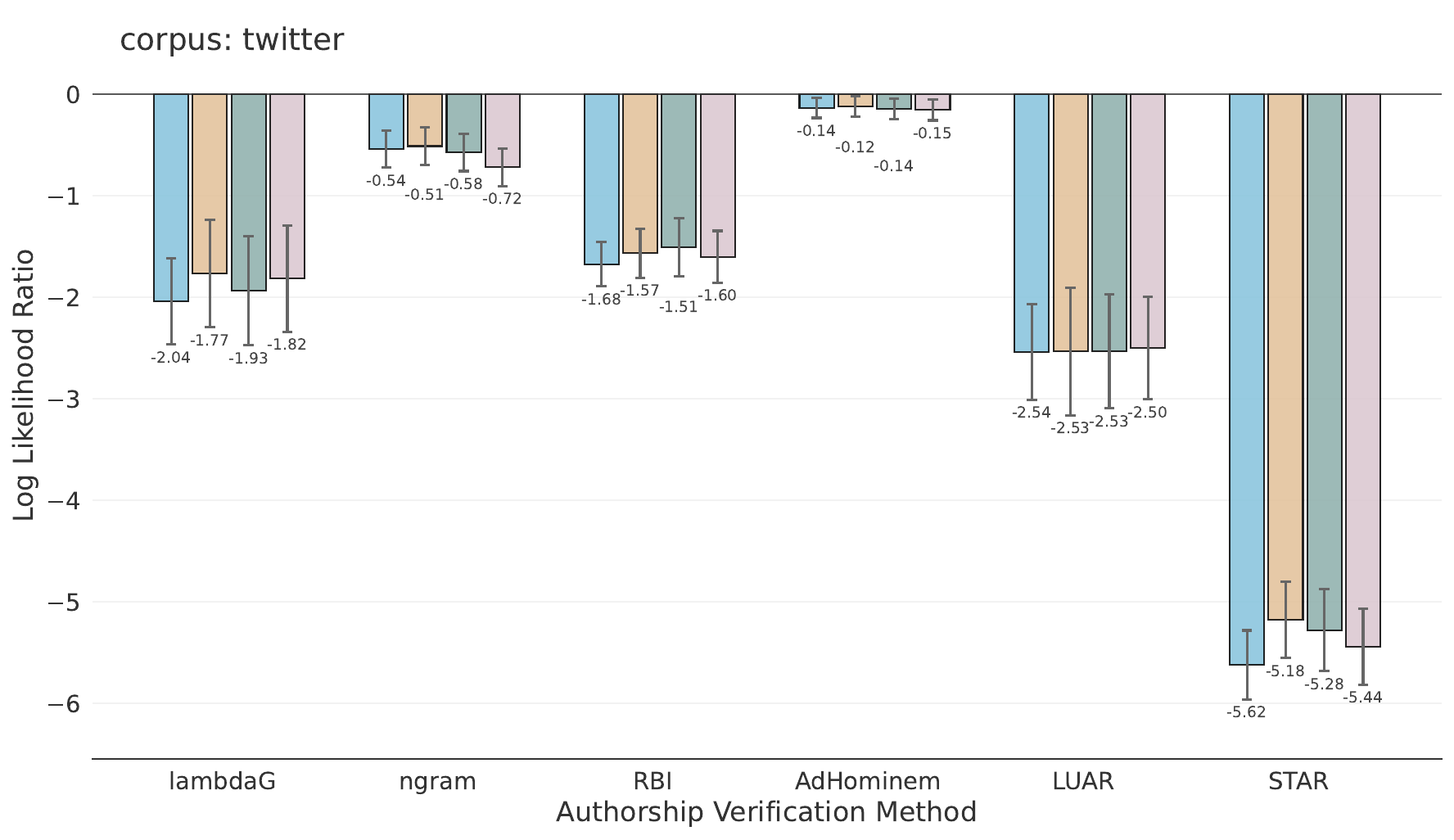}
     
    \caption{
    LLR scores produced by six authorship verification methods on LLM-generated impersonation texts under four prompting techniques, across three corpora: Enron (top), BOLT (middle), and Twitter (bottom). Bars indicate mean calibrated LLRs, with error bars representing 95\% confidence intervals. Results for non-neural authorship verification methods (LambdaG, n-gram tracing, and RBI) are obtained with POSNoise preprocessing applied, while neural network–based methods (AdHominem, LUAR, and STAR) are evaluated on texts without POSNoising.}
    \label{fig: LLR-separated}
\end{figure}

Figure~\ref{fig: LLR-separated} presents the performance of six AV methods on LLM-generated impersonation texts across four prompting techniques and three corpora. 
Results are reported as LLRs, where positive values indicate support for the same-author hypothesis, negative values indicate support for the different-author hypothesis, and larger absolute values reflect stronger evidential strength, with values close to zero indicating an inconclusive result.

Across all three corpora and across both non-neural and neural AV methods, a consistent overall pattern emerges: LLM-generated impersonations were largely unsuccessful. The vast majority of LLRs are negative, indicating that generated texts were generally rejected as having been written by the target author, regardless of the prompting strategy.

Despite this general resilience, method-specific variations are evident.
Among the neural approaches, STAR consistently produced strongly negative LLRs across all prompting conditions and all corpora, indicating a high degree of robustness to impersonation attempts. LUAR showed a similar pattern, though it yielded less definitive scores, indicating lower confidence in certain settings. 
In contrast, AdHominem was the most vulnerable, with scores clustering near zero, suggesting inconclusive judgments about authorship.

The non-neural methods also demonstrated overall robustness, though their relative behaviour differed by corpus. 
LambdaG exhibited broadly comparable performance across three corpora. By contrast, the other two methods, n-gram tracing and RBI, produced LLRs that fluctuated across datasets, reflecting varying levels of verification confidence. 
Despite these differences, all non-neural methods predominantly favoured the non-match hypothesis, confirming their resilience to LLM-generated texts.

Beyond method-specific behaviours, a clear corpus effect is evident across the evaluations. 
While LLRs remained predominantly negative across the board, the magnitude of these scores varied depending on the dataset. 
Specifically, the Enron corpus yielded LLRs with the smallest absolute values, indicating that AV methods were least confident in rejecting the impersonations within this email domain. 
Conversely, the Twitter corpus produced the most strongly negative LLRs across all prompting conditions, suggesting that LLM-generated imitations of social media texts were the easiest for AV systems to robustly identify and reject.

Variation across prompting techniques was limited. 
While minor fluctuations in LLRs can be observed for individual methods within specific corpora, no prompting strategy exhibited a consistent advantage across authorship verification methods or across datasets. 
Moreover, more structured prompting approaches, including planning-based and linguistically informed approaches did not yield systematically higher LLRs compared to naive prompting.

In sum, even with sophisticated prompting, GPT-4o failed to reproduce the fine-grained authorial markers that characterise an individual’s writing. 
Both non-neural and neural AV methods remained effective at distinguishing genuine texts from LLM-generated impersonations.

\subsection{TNR difference of AV Methods}

\begin{figure}[!b]
    \centering
    \includegraphics[width=0.98\linewidth]{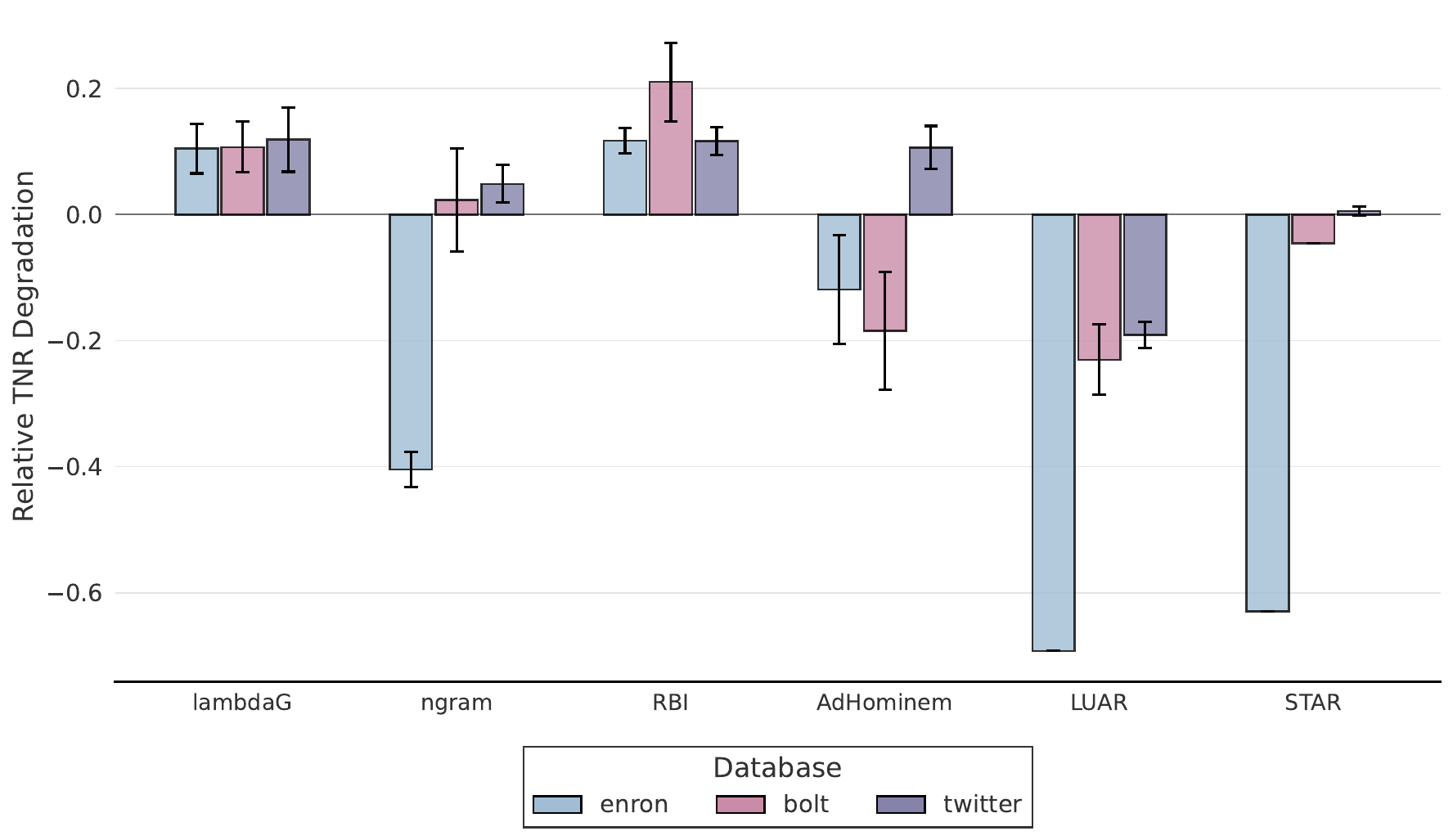}
    \includegraphics[width=0.98\linewidth]{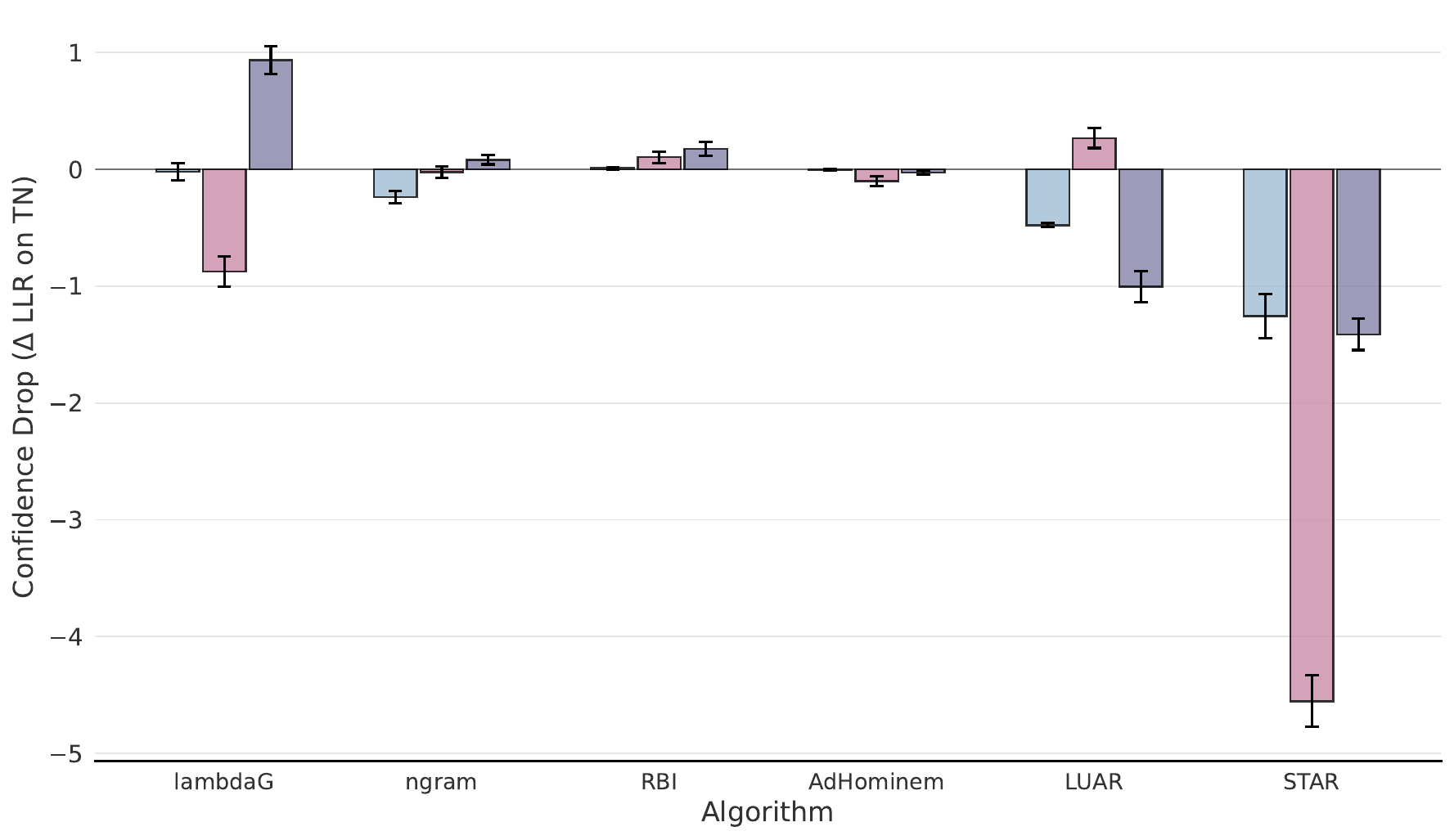}
    \caption{
    Performance degradation of authorship verification methods on LLM-generated impersonation texts compared to genuine test cases, averaged across four prompting conditions (naive, self-prompting, role-play, and tree-of-thoughts). Error bars represent 95\% confidence intervals.
    \textit{Top:} 
    TNR Degradation.
    Values are computed as $(\text{TNR}_{\text{test}} - \text{TNR}_{\text{impersonation}}) / \text{TNR}_{\text{test}}$.  
    Positive values represent the proportional loss of a method's original ability to correctly reject different-author pairs when facing generated impersonations, whereas negative values indicate \textit{improved} performance when rejecting LLM-generated impersonations.
    \textit{Bottom:} 
    Confidence drop on TN cases, measured as the difference in averaged LLR magnitudes ($|\text{LLR}_{\text{test}}| - |\text{LLR}_{\text{impersonation}}|$). 
    Since LLRs for true negative cases are inherently negative, this absolute difference quantifies the loss of evidential strength. 
    Positive values indicate that a method becomes less confident (i.e., assigns weaker evidential support to the different-author hypothesis) when rejecting LLM-generated impersonations, whereas negative values would indicate increased confidence.
    }
    \label{fig: TNR drop}
\end{figure}

To more directly assess whether AV methods were ``fooled'' by LLM-generated impersonations, we compared each algorithm’s performance on genuine corpus test cases with its performance on the impersonation texts. 
Since all impersonation texts form negative cases (i.e., not written by the claimed author), we report Relative True Negative Rate (TNR) Degradation. 
This metric calculates the proportional loss of a method's baseline ability to correctly reject different-author pairs, defined as $(\text{TNR}_{\text{test}} - \text{TNR}_{\text{impersonation}}) / \text{TNR}_{\text{test}}$.
This comparison controls for differences in corpus difficulty and method strength, allowing us to evaluate each system’s vulnerability to impersonations regardless of its baseline performance.

In addition to TNR, we further examined the confidence drop over True Negative (TN) cases. 
We include this measure to assess the change in the strength of the evidential support assigned by AV systems, or whether they become more hesitant or more certain when rejecting LLM-generated texts. 
This is computed as the difference in absolute magnitudes, defined as $|\text{LLR}_{\text{test}}| - |\text{LLR}_{\text{impersonation}}|$.
Given that earlier analyses revealed no prompting technique to consistently outperform others across corpora or methods, results are reported as averages across all four prompting conditions. 
Results are shown in Figure~\ref{fig: TNR drop}.

With respect to relative TNR degradation, LambdaG and RBI exhibit small but consistent decreases across all three corpora, indicating that these methods were at least occasionally misled by impersonation texts. 
In contrast, the remaining methods show more varied behaviour. 
Notably, STAR and LUAR reliably achieve higher TNRs on impersonation texts than on genuine human-authored test sets. This effect is particularly extreme on the Enron corpus, where LUAR's rejection rate improved by 69\% relative to its baseline, and n-gram tracing improved by 40\%. 
This indicates that these methods rejected LLM-generated impersonations more reliably than genuine negative cases.

Turning to the confidence drop (LLR differences) over TN cases, most AV methods only exhibit small differences between test data and impersonation texts. 
This suggests that, when correct decisions are made, the strength of the evidence assigned by these methods remains broadly comparable. 
An important exception is STAR, which consistently assigns stronger evidential support to the different-author hypothesis when rejecting LLM-generated impersonations across corpora. 
LUAR has a similar improvement in evidential strength for the Enron corpus, but this pattern does not generalise to the BOLT and Twitter datasets.
Ultimately, these findings reveal that state-of-the-art neural-based AV methods are not only rarely deceived by LLM impersonations, but they often reject them with greater confidence.
We return to this surprising pattern in Section~\ref{sec: diversity}.

\section{Discussion}
\label{sec: discussion}

The present study aimed to examine whether prompted LLMs can generate convincing impersonations of human authors, and whether state-of-the-art AV methods are robust against such attempts. 
By evaluating multiple prompting strategies across range of non-neural and neural verification algorithms, our results offer a comprehensive empirical perspective on the emerging forensic challenge posed by LLM-assisted authorship impersonation.

\subsection{Impersonation Capabilities of Prompted LLMs}

First, our findings indicate that LLMs remain limited in their ability to capture the linguistic markers that define a human author's individuality via prompting alone. 
Even with access to genuine samples provided in the prompt, GPT-4o failed to generate passages that AV systems would consistently judge as being authored by the same individual.
This suggests that, under a few-shot in-context learning setting, LLMs struggle to reproduce the idiolectal features that characterise a specific author’s linguistic identity.

These results are broadly consistent with previous studies reporting that LLMs tend to reproduce surface-level features rather than deeper linguistic individuality~\cite{bhandarkar2024emulating, chen2024using}. 
However, direct comparison is complicated by differences in genre, datasets and evaluation frameworks.
For example, while Wang et al.~\cite{wang2025catch} observed that LLMs perform relatively well at impersonating structured genres such as emails, their evaluation relies on a custom neural AV model rather than established forensic verification methods. 
In contrast, our results provide a stronger and more rigorous and operationally relevant assessment by evaluating LLM outputs against validated verification systems.

Second, the evaluation of authorship verification methods shows that both non-neural and neural–based models retained strong discriminative power against LLM-generated texts. 
Among neural models, AdHominem frequently produced scores close to zero, yielding comparatively indecisive judgments. 
In contrast, LUAR and STAR consistently demonstrated robust behaviour across corpora, with both methods achieving higher true negative rates on impersonation texts than on genuine test data. 
On the non-neural side, n-gram tracing, the RBI method, and LambdaG all showed substantial resilience. 
Although LambdaG and RBI exhibited small reductions in TNR when evaluating impersonation texts, these decreases were limited in magnitude and did not undermine their overall ability to reject non-matching authorship.
This finding reinforces the value of simple, interpretable baselines in forensic contexts.

Third, the limited variation observed across prompting techniques suggests that prompt engineering alone is insufficient to achieve reliable authorship impersonation.
Although different prompting strategies, ranging from naive rewriting to more structured, planning-based approaches, produced minor performance fluctuations, none of them enabled the LLM to consistently bypass AV systems across methods or corpora.
This finding aligns with observations from previous studies~\cite{chen2024using, chatgptReview_thompson2025unveiling}, pointing to a broader limitation: the model’s capacity for few-shot in-context imitation does not extend to replicating the stable, author-specific markers that AV methods rely on.

Finally, our results reveal clear dataset-dependent differences in evidential strength. 
Across all analyses, impersonation texts in the Twitter corpus yielded the most strongly negative LLRs, whereas Enron emails produced smaller absolute LLR magnitudes, indicating lower confidence in the different-author judgment. 
This suggests that the inherent characteristics of different corpora play an important role in shaping the evidential strength assigned by authorship verification systems.

\subsection{Lexical Diversity of LLM-generated Text}
\label{sec: diversity}

When comparing TNRs between genuine corpus test sets and LLM-generated impersonation texts, we observed a notable contrast across methods. 
While several AV methods exhibited modest performance degradation on impersonation data, STAR and LUAR consistently showed improved TNR across all three corpora, indicating that these models rejected impersonation texts more reliably than genuine negative cases. 
Additionally, n-gram tracing demonstrated a marked improvement on the Enron corpus. 
Beyond binary decision accuracy, STAR further assigned stronger evidential support to correct rejections, as reflected in higher average LLR magnitudes over TN cases. 
Together, these patterns suggest that certain methods are not only robust to impersonation attempts, but may in fact exploit systematic properties of LLM-generated texts.

\begin{figure}[!b]
    \centering
    \includegraphics[width=0.98\linewidth]{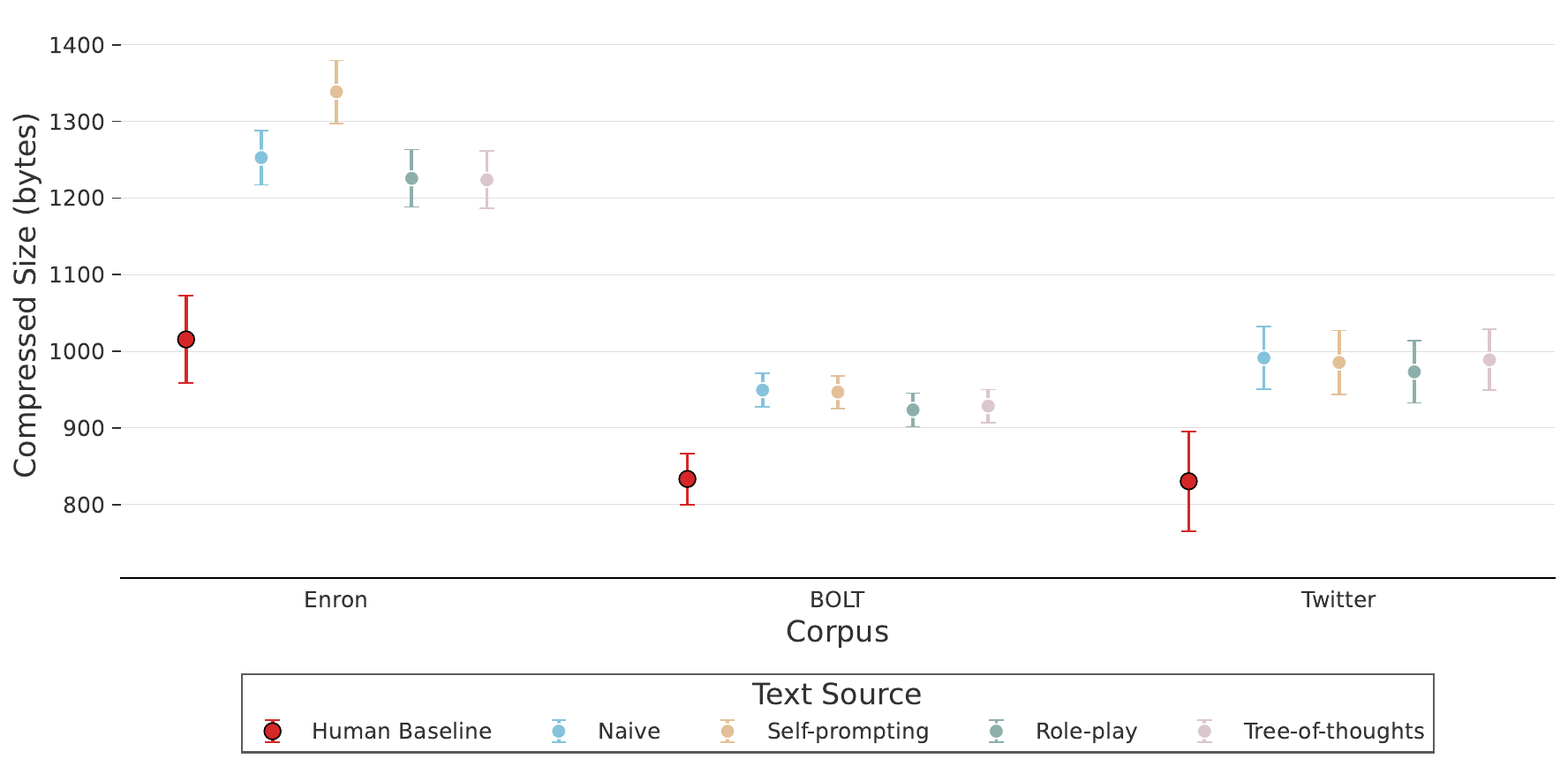}
    \includegraphics[width=0.98\linewidth]{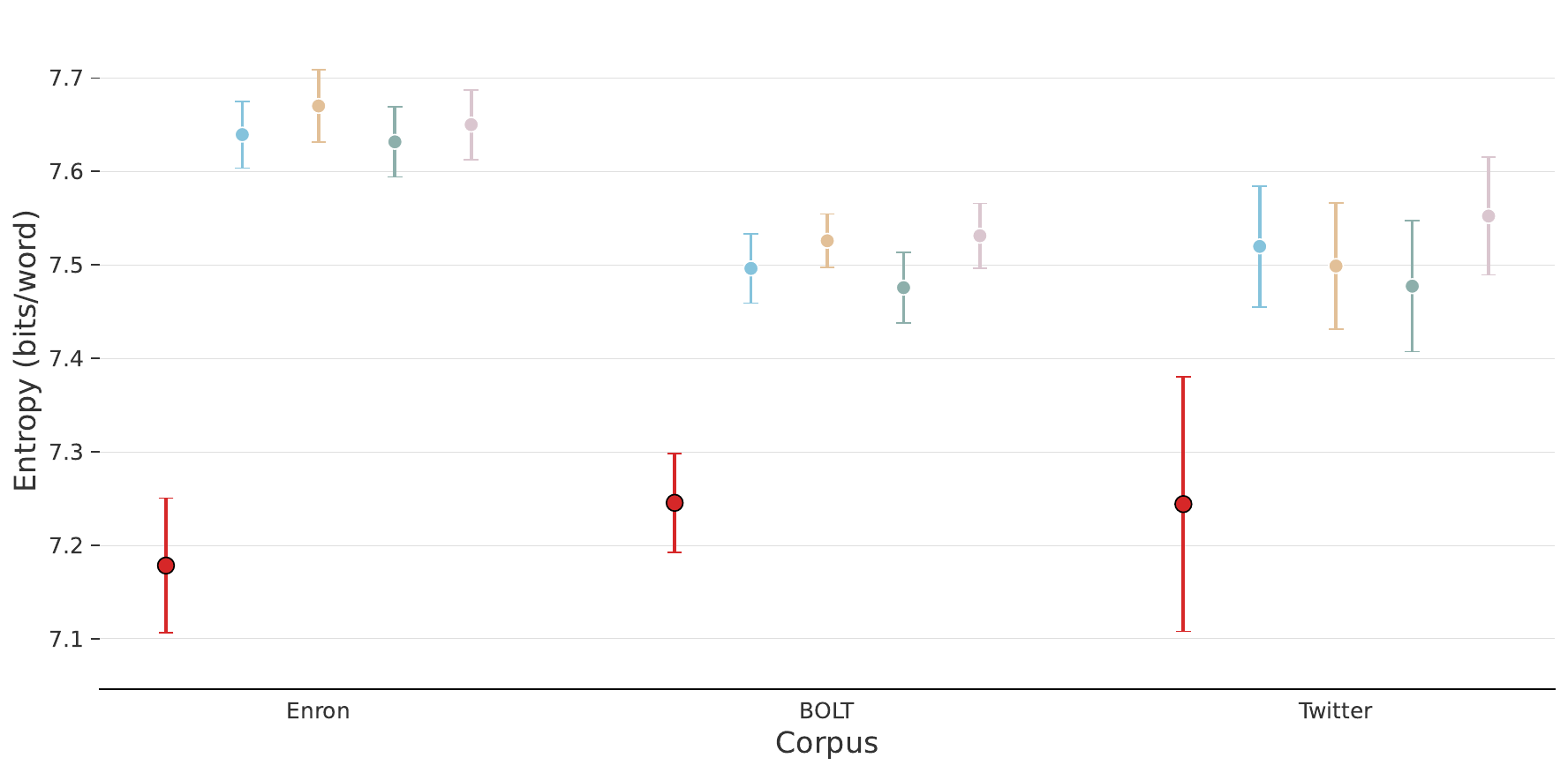}
    \includegraphics[width=0.98\linewidth]{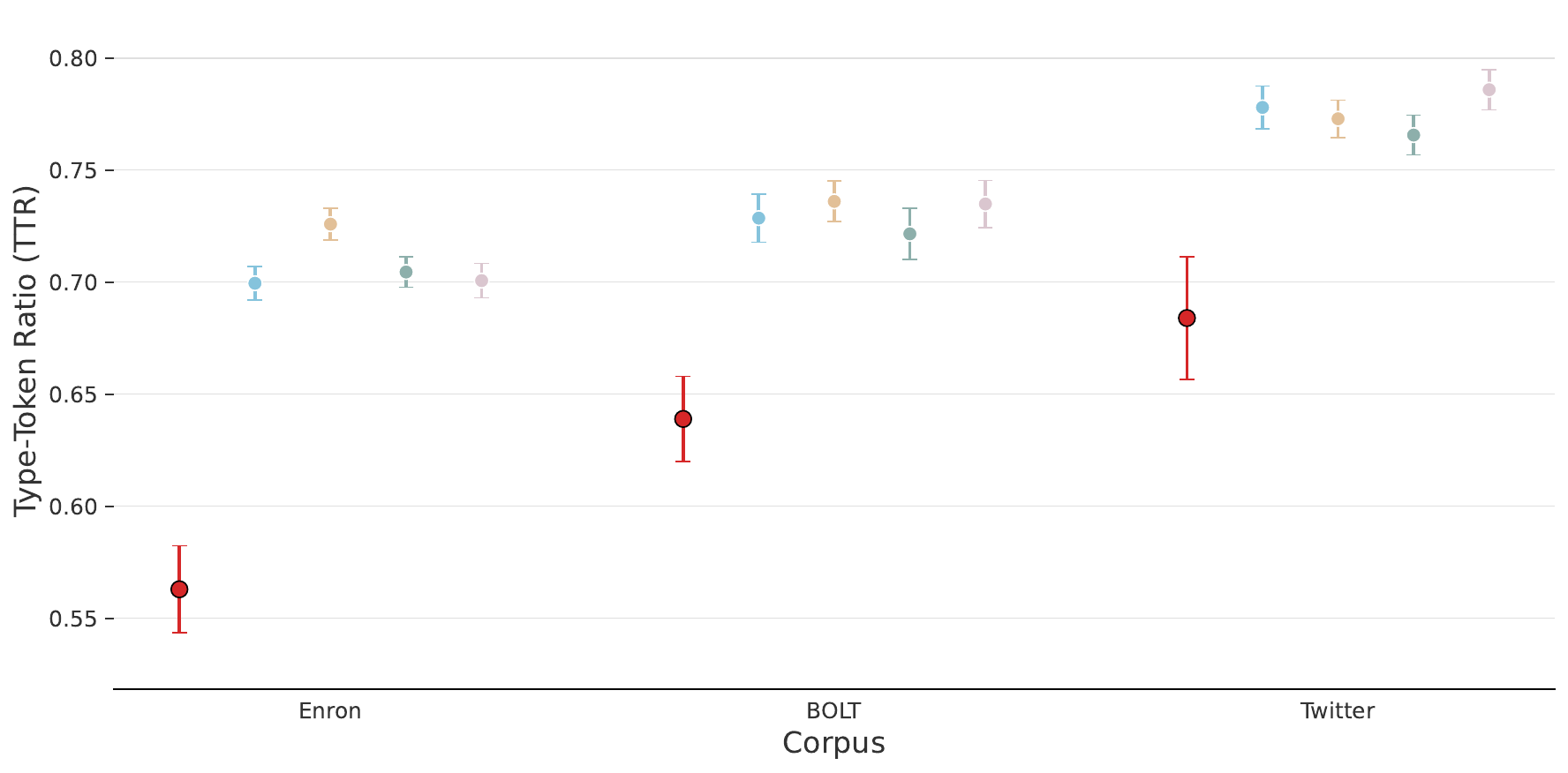}
    \caption{
    Comparison between LLM-generated impersonation texts and human-authored texts in terms of compressed size (top), entropy (middle), and TTR (bottom). 
    Error bars indicate 95\% confidence intervals.
    Note that the y-axes are scaled to the specific data range of each metric.
    }
    \label{fig: follow-up}
\end{figure}

To explore this counter-intuitive finding, we conducted a follow-up analysis comparing the lexical diversity of human-authored texts with that of LLM-generated impersonations. 
We applied three complementary measures:
a) Compressed size: the number of bytes after compression, which reflects textual redundancy and repetitiveness in the text; 
b) Entropy: a measure of distributional unpredictability across tokens; and
c) Type–Token Ratio (TTR): the proportion of unique word types relative to total tokens. 
For these analyses, all texts were truncated to the same length prior to calculation to ensure comparability.

As shown in Figure~\ref{fig: follow-up}, LLM-generated texts consistently exhibited larger compressed size, higher entropy, and higher TTRs than samples from the target author across all three corpora. 
This pattern indicates that generated texts are systematically more lexically diverse and less redundant than authentic human writing, regardless of genre or prompting strategy. 
Such properties are atypical of individual writing idiolects, which are inherently characterised by recurring lexical choices, preferred constructions, and predictable redundancies. 

These cross-corpus regularities provide a plausible explanation for the observed robustness of certain authorship verification methods. 
We first consider n-gram tracing, whose underlying mechanism is highly interpretable. 
The improved performance of this method, particularly on the Enron corpus, directly aligns with our findings on lexical diversity. 
Human-authored texts naturally contain recurring character or word n-grams that reflect an individual's habitual phrasing. 
Because LLM-generated texts exhibit higher entropy and greater lexical diversity, they fail to produce sufficient repetitions of these local sequences, which makes the impersonations stand out.

For the neural network-based methods, STAR and LUAR, the internal representations are not directly interpretable. 
Given the differences between human and LLM writing, it is possible that these models are inadvertently performing a form of AI text detection alongside authorship verification. 
Because STAR and LUAR are trained to capture deep, style-sensitive representations, they might be highly sensitive to the overarching ``machine-like'' signature of the text. 
In other words, when presented with an LLM-generated impersonation, these models might reject it not simply because it lacks the specific target author's language, but because its fundamental characteristics (such as inflated diversity) deviate so significantly from natural human writing.

It should also be noted that in our text generation setup, decoding parameters such as temperature and frequency penalty were kept at their default values. 
Since these parameters directly affect lexical diversity and entropy, future work could systematically vary them to examine whether controlling for output randomness influences the detectability of LLM-generated impersonations.

\subsection{Limitations and Future Work}

This study has several limitations that open avenues for future research. 
First, our experiments were designed to simulate an entry-level malicious actor who relies solely on straightforward prompting strategies. 
More sophisticated attackers might adopt adversarial tactics, such as using feedback from AV systems themselves to iteratively refine prompts and converge on more convincing impersonations. 
Exploring such adaptive strategies will be critical to fully assess the resilience of AV methods under realistic adversarial conditions. 
In addition, future work could investigate the impact of parameter-efficient fine-tuning (PEFT) techniques, which may enable attackers to customise models toward a specific authorial profile with minimal data and computational cost.

Second, our evaluation framework did not incorporate human judgment. 
While AV algorithms reliably rejected most LLM-generated impersonations, some passages may still appear convincing to human readers. 
In forensic contexts, this human susceptibility is highly relevant: impersonated messages may deceive family members, colleagues, or investigators, even if algorithms can successfully detect them. 
Future work could therefore integrate human-subject studies to assess the perceived authenticity of LLM-generated impersonations alongside or in contrast to computational measures.

\section{Conclusion}
\label{sec: conclusion}

This study investigated whether LLMs can be prompted to impersonate individual authors and whether the resulting adversarially generated texts can bypass state-of-the-art AV systems. 
Using three forensically relevant corpora and a range of prompting strategies, we found that GPT-4o was unable to convincingly replicate the stable linguistic markers that define authorial individuality. 
Both non-neural and neural AV approaches largely resisted these impersonation attempts. 
These findings indicate a continued robustness of current AV systems.

Despite these promising results, our findings highlight several limitations and directions for future research. 
Certain neural architectures, most notably AdHominem, exhibited partial vulnerability, suggesting that specific model designs may be more susceptible to adversarially generated text. 
Moreover, while our experiments focused on entry-level adversaries relying on straightforward prompting techniques, more sophisticated attackers, such as those employing adaptive strategies or model customisation, may pose a greater challenge to forensic analysis. 
Finally, although AV algorithms consistently rejected LLM-generated texts, such texts may still appear plausible to human readers, raising concerns about real-world deception.

Overall, by situating LLM-based authorship impersonation within a forensic science framework and systematically evaluating its interaction with both non-neural and neural AV methods, this study provides empirical evidence that current verification systems remain effective against prompted impersonation attacks under realistic conditions.

%% file: appendix.tex
\appendix
\subsection{Prompts used in experiments}
\label{app:prompts}

\subsubsection*{Naive prompting}
Rewrite the given original text so that it appears to have been written by the author of the provided example text snippets. 
\textit{\{Output instruction:\}} As output, exclusively return the rewritten text without any accompanying explanations or comments.
\textit{\{original text\} + \{concatenated example snippets\}}.

\subsubsection*{Self-prompting}
\textit{Meta-prompt:} Write a prompt to make a large language model impersonate a human author. The goal is to prompt the model to rewrite a given text as if it had been written by the same author as the provided text snippets.
Write an effective prompt for the LLM to perform this task.  
\textit{\{Output instruction\}} + 
\textit{\{original text\} + \{concatenated example snippets\}}.

\subsubsection*{Role-playing prompting}
\paragraph{Role conditioning}
\textit{(Given in the developer prompt)} You are a writing assistant that specializes in mimicking the unique linguistic characteristics of human authors. 
\paragraph{Detailed instruction}
\textit{(Given in the user prompt)} Your goal is to rewrite the input text as if it had been written by a specific author, using a set of example snippets to guide the adaptation. 
Maintain the meaning and intent of the original message while making it read as though it were produced by the same author of the examples.
Disregard the differences in topic and content of the example snippets. Pay close attention to linguistic features such as phrasal verbs, modal verbs, punctuation, rare words, affixes, quantities, humour, sarcasm, typographical errors, and misspellings. 
\textit{\{Output instruction\}} + 
\textit{\{original text\} + \{concatenated example snippets\}}.

\subsubsection*{ToT prompting}
You are a writing assistant that specialises in mimicking the unique linguistic characteristics of human authors.
Your goal is to rewrite the input text as if it had been written by a specific author, using a set of example snippets to guide the adaptation. 
Maintain the meaning and intent of the original message while making it read as though it were produced by the same author of the examples.
Disregard the differences in topic and content of the example snippets. Pay close attention to linguistic features such as phrasal verbs, modal verbs, punctuation, rare words, affixes, quantities, humour, sarcasm, typographical errors, and misspellings.  
\textit{\{original text\} + \{concatenated example snippets\}}.

\textit{For plan generation:}
You have to make a plan of how to achieve this goal. 
\textit{\{Output instruction\}} 

\textit{For text generation:}
You have to follow the given plan to achieve this goal.
\textit{\{Output instruction\}} 

\textit{For voting:}
Above is the instruction for a task, analyse choices below, then conclude which is most promising for the instruction. 
\textit{\{Output instruction\}} 